\def\year{2020}\relax
\documentclass[letterpaper]{article} 
\usepackage{aaai20}  
\usepackage{times}  
\usepackage{helvet} 
\usepackage{courier}  
\usepackage[hyphens]{url}  
\usepackage{graphicx} 
\usepackage{graphicx}  
\usepackage{arydshln}
\usepackage{amsmath}
\usepackage{multirow} 
\usepackage{multicol}
\usepackage{pgfplots}
\DeclareMathOperator*{\argmax}{argmax}
\usepackage{algorithm}
\usepackage{algorithmic}
\usepackage{booktabs}
\title{Global Greedy Dependency Parsing}

\author{Zuchao Li$^{1,2,3}$, Hai Zhao$^{1,2,3,}$\thanks{$\ $ Corresponding author. This paper was partially supported by National Key Research and Development Program of China (No. 2017YFB0304100) and Key Projects of National Natural Science Foundation of China (No. U1836222 and No. 61733011).}, 	Kevin Parnow$^{1,2,3}$\\
	$^{1}$Department of Computer Science and Engineering, Shanghai Jiao Tong University \\
	$^{2}$Key Laboratory of Shanghai Education Commission for Intelligent Interaction \\ and Cognitive Engineering, Shanghai Jiao Tong University, Shanghai, China\\
	$^{3}$MoE Key Lab of Artificial Intelligence, AI Institute, Shanghai Jiao Tong University \\
	{\tt charlee@sjtu.edu.cn, zhaohai@cs.sjtu.edu.cn, parnow@sjtu.edu.cn}
}

\begin{document}

\maketitle

\begin{abstract}

    Most syntactic dependency parsing models may fall into one of two categories: transition- and graph-based models. The former models enjoy high inference efficiency with linear time complexity, but they rely on the stacking or re-ranking of partially-built parse trees to build a complete parse tree and are stuck with slower training for the necessity of dynamic oracle training. The latter, graph-based models, may boast better performance but are unfortunately marred by polynomial time inference. 
    In this paper, we propose a novel parsing order objective, resulting in a novel dependency parsing model capable of both global (in sentence scope) feature extraction as in graph models and linear time inference as in transitional models. The proposed global greedy parser only uses two arc-building actions, left and right arcs, for projective parsing. When equipped with two extra non-projective arc-building actions, the proposed parser may also smoothly support non-projective parsing.
    Using multiple benchmark treebanks, including the Penn Treebank (PTB), the CoNLL-X treebanks, and the Universal Dependency Treebanks, we evaluate our parser and demonstrate that the proposed novel parser achieves good performance with faster training and decoding.
    
\end{abstract}

\section{Introduction}

\begin{figure}
	\centering
	\includegraphics[width=0.47\textwidth]{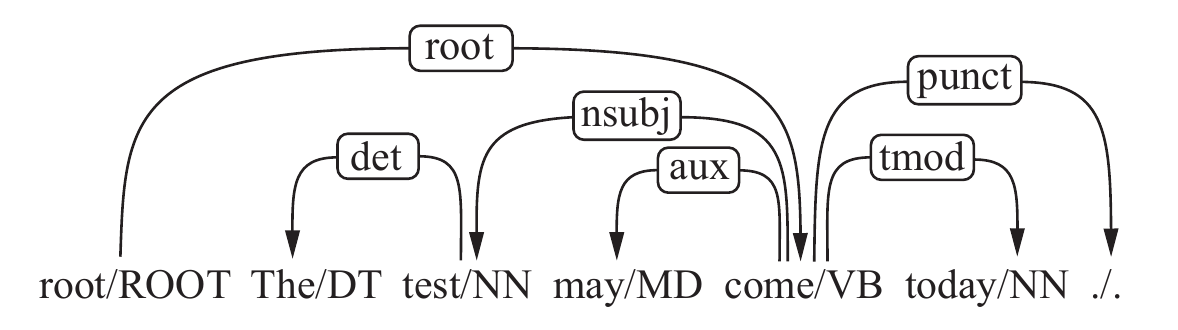}
	\caption{A fully built dependency tree for ``\emph{The test may come today.}" including part-of-speech (POS) tags and \emph{root} token.} \label{dp_example}
\end{figure}

Dependency parsing predicts the existence and type of linguistic dependency relations between words (as shown in Figure \ref{dp_example}), which is a critical step in accomplishing deep natural language processing. Dependency parsing has been well developed \cite{Stephen2011A,li-etal-2018-seq2seq}, and it generally relies on two types of parsing models: transition-based models and graph-based models. The former \cite{yamada2003statistical,nivre2004deterministic,zhang2011transition,Zhao2009Cross,zhang2011transition} traditionally apply local and greedy transition-based algorithms, while the latter \cite{eisner1996three,mcdonald2005online,mcdonald2005non,ma-zhao-2012-fourth,zhang-etal-2016-probabilistic,li-etal-2018-joint-learning,li2019cross} apply globally optimized graph-based algorithms. 


A transition-based dependency parser processes the sentence word-by-word, commonly from left to right, and forms a dependency tree incrementally from the operations predicted. This method is advantageous in that inference on the projective dependency tree is linear in time complexity with respect to sentence length; however, it has several obvious disadvantages. Because the decision-making of each step is based on partially-built parse trees, special training methods are required, which results in slow training and error propagation, as well as weak long-distance dependence processing \cite{mcdonald2011analyzing}.


Graph-based parsers learn scoring functions in one-shot and then perform an exhaustive search over the entire tree space for the highest-scoring tree. This improves the performances of the parsers, particularly the long-distance dependency processing, but these models usually have slow inference speed to encourage higher accuracy. 

The easy-first parsing approach \cite{kiperwasser2016easy,LiEffective} was designed to integrate the advantages of graph-based parsers’ better-performing trees and transition-based parsers’ linear decoding complexity. By processing the input tokens in a stepwise easy-to-hard order, the algorithm makes use of structured information on partially-built parse trees. Because of the presence of rich, structured information, exhaustive inference is not an optimal solution - we can leverage this information to conduct inference much more quickly. As an alternative to exhaustive inference, easy-first chooses to use an approximated greedy search that only explores a tiny fraction of the search space. Compared to graph-based parsers, however, easy-first parsers have two apparent weaknesses: slower training and worse performance. According to our preliminary studies, with the current state-of-the-art systems, we must either sacrifice training complexity for decoding speed, or sacrifice decoding speed for higher accuracy. 


In this paper, we propose a novel Global (featuring) Greedy (inference) parsing architecture\footnote{Our code is available at \url{https://github.com/bcmi220/ggdp}.} that achieves fast training, high decoding speed and good performance. With our approach, we use the one-shot arc scoring scheme as in the graph-based parser instead of the stepwise local scoring in transition-based. This is essential for achieving competitive performance, efficient training, and fast decoding. Since, to preserve linear time decoding, we chose a greedy algorithm, we introduce a parsing order scoring scheme to retain the decoding order in inference to achieve the highest accuracy possible. Just as with one-shot scoring in graph-based parsers, our proposed parser will perform arc-attachment scoring, parsing order scoring, and decoding simultaneously in an incremental, deterministic fashion just as transition-based parsers do.




We evaluated our models on the common benchmark treebanks PTB and CTB, as well as on the multilingual CoNLL and the Universal Dependency treebanks. From the evaluation results on the benchmark treebanks, our proposed model gives significant improvements when compared to the baseline parser. In summary, our contributions are thus:


$\bullet$ We integrate the arc scoring mechanism of graph-based parsers and the linear time complexity inference approach of transition parsing models, which, by replacing stepwise local feature scoring, significantly alleviates the drawbacks of these models, improving their moderate performance caused by error propagation and increasing their training speeds resulting from their lack of parallelism.




$\bullet$  Empirical evaluations on benchmark and multilingual treebanks show that our method achieves state-of-the-art or comparable performance, indicating that our novel neural network architecture for dependency parsing is simple, effective, and efficient.


$\bullet$ Our work shows that using neural networks’ excellent learning ability, we can simultaneously achieve both improved accuracy and speed.


\section{The General Greedy Parsing}




The global greedy parser will build its dependency trees in a stepwise manner without backtracking, which takes a general greedy decoding algorithm as in easy-first parsers.




Using easy-first parsing's notation, we describe the decoding in our global greedy parsing. As both easy-first and global greedy parsing rely on a series of deterministic parsing actions in a general parsing order (unlike the fixed left-to-right order of standard transitional parsers), they need a specific data structure which consists of a list of unattached nodes (including their partial structures) referred to as ``\emph{pending}". At each step, the parser chooses a specific action $\hat{a}$ on position $i$ with the given arc score \emph{score}($\cdot$), which is generated by an arc scorer in the parser. Given an intermediate state of parsing with \emph{pending} $P=\{p_0, p_1, p_2, \cdots, p_N\}$, the attachment action is determined as follows:
$$\hat{a} = \argmax\limits_{\substack{act \in \mathcal{A},\ 1\leq i \leq N}} \ score(act(p_i)),$$
where $\mathcal{A}$ denotes the set of the allowed actions, and $i$ is the index of the node in \emph{pending}. In addition to distinguishing the correct attachments from the incorrect ones, 
the arc scorer also assigns the highest scores to the easiest attachment decisions and lower scores to the harder decisions, thus determining the parsing order of an input sentence.


For projective parsing, there are exactly two types of actions in the allowed action set: \text{\large{A}}TTACH\text{\large{L}}EFT($i$) and \text{\large{A}}TTACH\text{\large{R}}IGHT($i$). Let $p_i$ refer to $i$-th element in \emph{pending}, then the allowed actions can be formally defined as follows:

$\bullet$ \text{\large{A}}TTACH\text{\large{L}}EFT($i$): attaches $p_{i+1}$ to $p_i$ , which results in an arc ($p_i$, $p_{i+1}$) headed by $p_i$, and removes $p_{i+1}$ from \emph{pending}.

$\bullet$ \text{\large{A}}TTACH\text{\large{R}}IGHT($i$): attaches $p_i$ to $p_{i+1}$ , which results in an arc ($p_{i+1}$, $p_i$) headed by $p_{i+1}$, and removes $p_i$ from \emph{pending}.


\begin{figure*}
	\centering
	\includegraphics[width=0.9\textwidth]{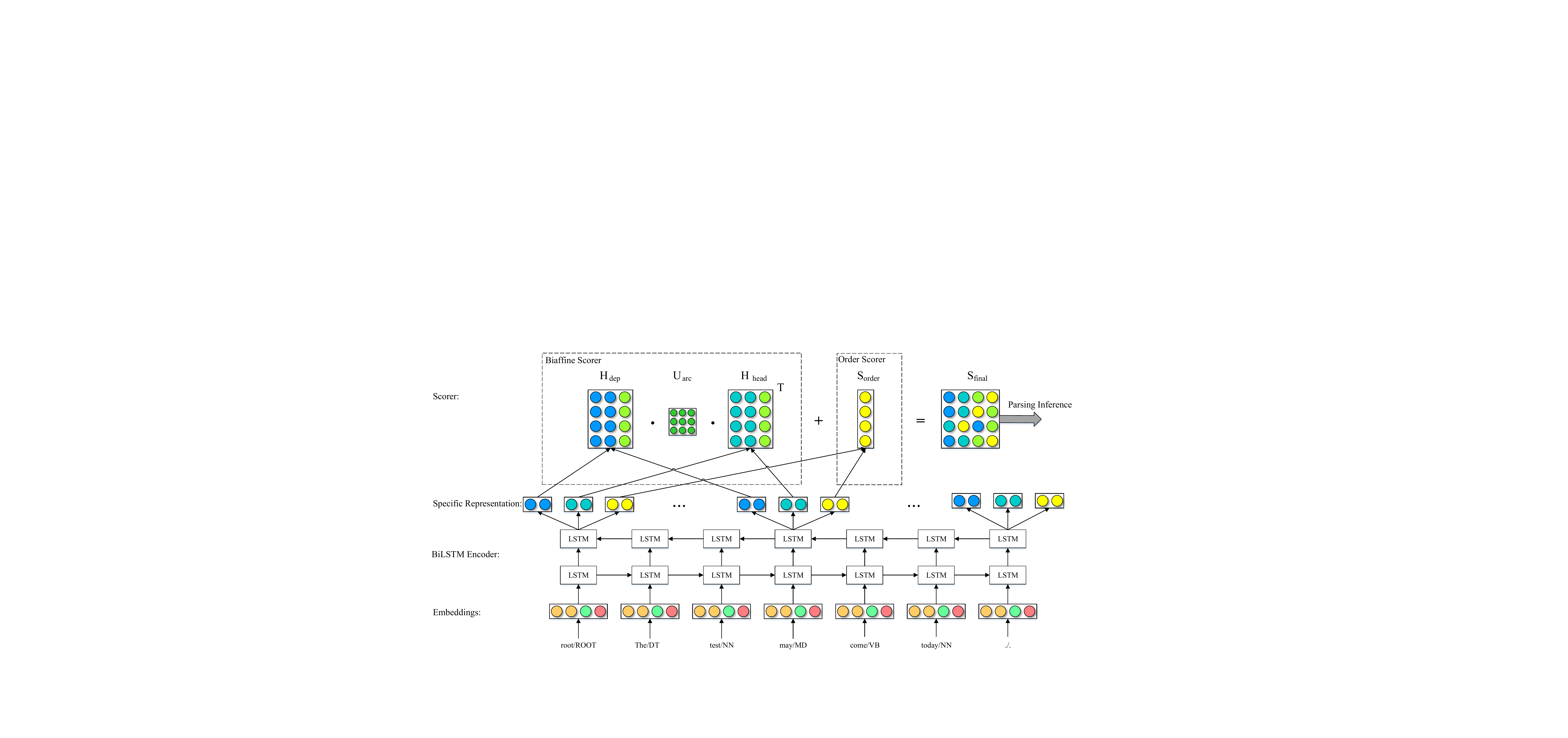}
	\caption{Overview for our global greedy parser.} \label{overview}
\end{figure*}

\section{Global Greedy Parsing Model}

Our proposed global greedy model contains three components: (1) an encoder that processes the input sentence and maps it into hidden states that lie in a low dimensional vector space $h_i$ and feeds it into a specific representation layer to strip away irrelevant information, (2) a modified scorer with a parsing order objective, and (3) a greedy inference module that generates the dependency tree.

\subsection{Encoder}

We employ a bi-directional LSTM-CNN architecture (BiLSTM-CNN) to encode the context in which convolutional neural networks (CNNs) learn character-level information $e_{char}$ to better handle out-of-vocabulary words. We then combine these words' character level embeddings with their word embedding $e_{word}$ and POS embedding $e_{pos}$ to create a context-independent representation, which we then feed into the BiLSTM to create word-level context-dependent representations. To further enhance the word-level representation, we leverage an external fixed representation $e_{lm}$ from pre-trained ELMo  \cite{peters2018deep} or BERT \cite{devlin2018bert} layer features. Finally, the encoder outputs a sequence of contextualized representations $h_i$.
$$h_{i}= BiLSTM([e^i_{word}; e^i_{pos}; e^i_{char}; e^i_{lm}])$$

Because the contextualized representations will be used for several different purposes in the following scorers, it is necessary to specify a representation for each purpose. As shown in \cite{dozat2017deep}, applying a multi-layer perceptron (MLP) to the recurrent output states before the classifier strips away irrelevant information for the current decision, reducing both the dimensionality and the risk of model overfitting. Therefore, in order to distinguish the biaffine scorer's head and dependent representations and the parsing order scorer's representations, we add a separate contextualized representation layer with ReLU as its activation function for each syntax head $h^{head}_i \in H_{head}$ specific representations, dependent $h^{dep}_i \in H_{dep}$ specific representations, and parsing order $h^{order}_i \in H_{order}$:
\begin{align*}
	h^{\it{m}}_i = ReLU(MLP_{\it{m}}(h_{i})), \it{m} \in [head, dep, order]
\end{align*}

\subsection{Scorers}

The traditional easy-first model relies on an incremental tree scoring process with stepwise loss backpropagation and sub-tree removal facilitated by local scoring, relying on the scorer and loss backpropagation to hopefully obtain the parsing order. Communicating the information from the scorer and the loss requires training a dynamic oracle, which exposes the model to the configurations resulting from erroneous decisions. This training process is done at the token level, not the sentence level, which unfortunately means incremental scoring prevents parallelized training and causes error propagation. We thus forego incremental local scoring, and, inspired by the design of graph-based parsing models, we instead choose to score all of the syntactic arc candidates in one-shot, which allows for global featuring at a sentence level; however, the introduction of one-shot scoring brings new problems. Since the graph-based method relies on a tree space search algorithm to find the tree with the highest score, parsing order is not important at all since graph-based inference. If we apply one-shot scoring to greedy parsing, we need a mechanism like a stack (as is used in transition-based parsing) to preserve the parsing order.

Both transition-based and easy-first parsers build parse trees in an incremental style, which forces tree formation to follow an order starting from either the root and working towards the leaf nodes or vice versa. When a parser builds an arc that skips any layer, certain errors will exist that it will be impossible for the parent node to find. We thus implement a parsing order prediction module to learn a parsing order objective that outputs a parsing order score addition to the arc score to ensure that each \emph{pending} node is attached to its parent only after all (or at least as many as possible) of its children have been collected.

\begin{figure}
	\centering
	\includegraphics[width=0.42\textwidth]{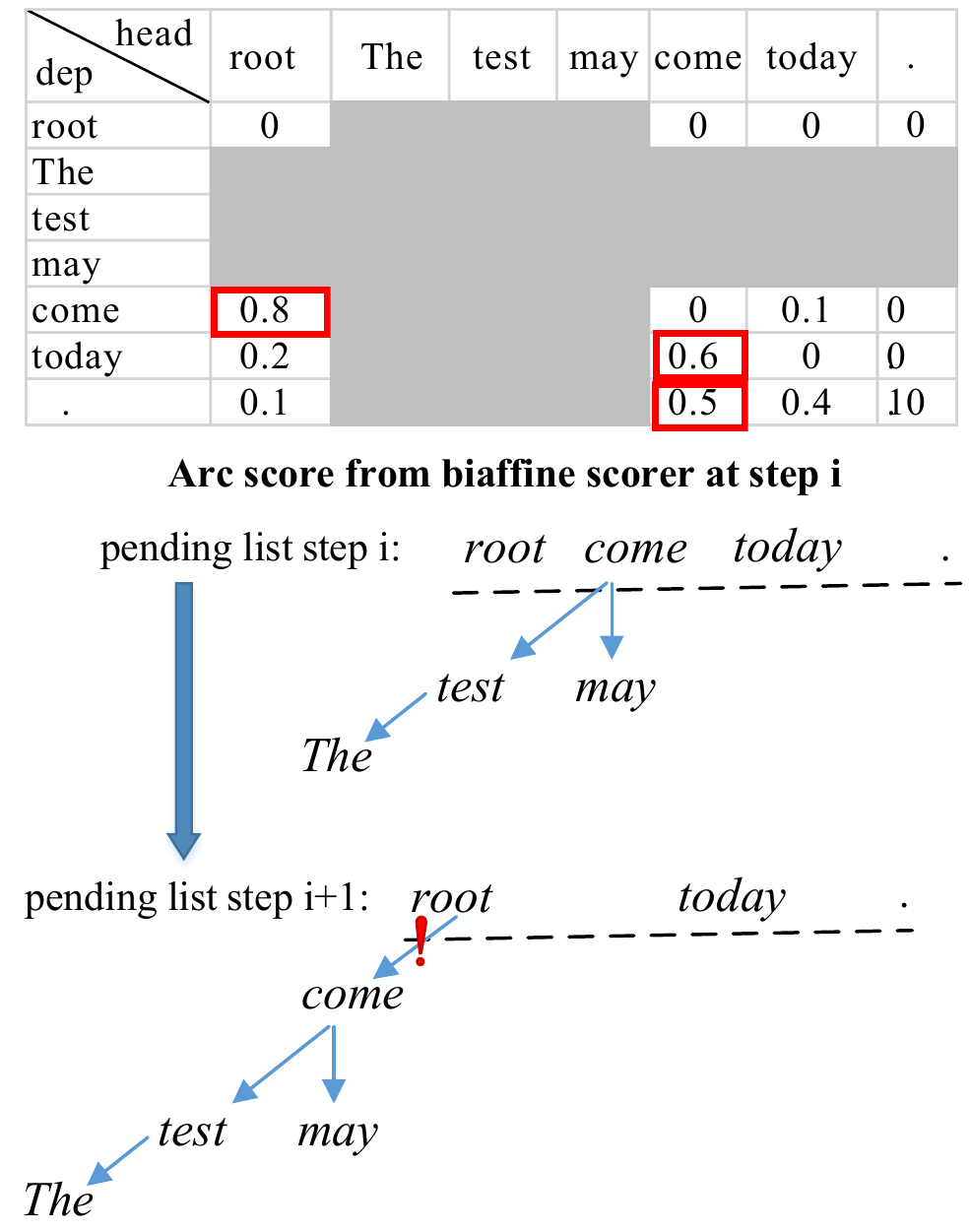}
	\caption{Illustration of the reason of errors generated when only using biaffine arc score for easy-first parsing.}\label{error}
\end{figure}

Our scorer consists of two parts: a biaffine scorer for one-shot scoring and a parsing order scorer for parsing order guiding. For the biaffine scorer, we adopt the biaffine attention mechanism \cite{dozat2017deep} to score all possible head-dependent pairs:
\begin{align*}
	s_{arc} &= H^T_{head} \textbf{W}_{arc} H_{dep} \\&+ \textbf{U}_{arc}^T H_{head} + \textbf{V}_{arc}^T H_{dep} + \textbf{b}_{arc}
\end{align*}
where $\textbf{W}_{arc}$, $\textbf{U}_{arc}$, $\textbf{V}_{arc}$, $\textbf{b}_{arc}$ are the weight matrix of the bi-linear term, the two weight vectors of the linear terms, and the bias vector, respectively.


If we perform greedy inference only on the $s_{arc}$ directly, as in Figure \ref{error}, at step $i$, the decoder tests every pair in the pending list, and although the current score fits the correct tree structure for this example, because backtracking is not allowed in the deterministic greedy inference, according to the maximum score $s_{arc}$, the edge selected in step $i$+$1$ is ``\textit{root}"$\rightarrow$``\textit{come}".  This prevents the child nodes (``\textit{today}" and ``\textit{.}") from finding the correct parent node in the subsequent step. Thus, the decoder is stuck with this error. This problem can be solved or mitigated by using a max spanning tree (MST) decoder or by adding beam search method to the inference, but neither guarantees maintaining linear time decoding. Therefore, we propose a new scorer for parsing order $s_{order}$. In the scoring stage, the parsing order score is passed to the decoder to guide it and prevent (as much as possible) resorting to erroneous choices.

We formally define the parsing order score for decoding. To decode the nodes at the bottom of the syntax tree first, we define the the parsing order priority as the layer ``level" or ``position" in the tree\footnote{``layer" is used to represent the distance between the current node to the ``root".}. The biaffine output score is the probability of edge (dependency) existence, between 0 and 1, so the greater the probability, the more likely an edge is to exist. Thus,  our parsing order scorer gives a layer score for a node, and then, we add this layer score to the biaffine score. Consequently, the relative score of the same layer can be kept unchanged, and the higher the score of a node in the bottom layer, the higher its decoding priority will be. We therefore define $s_{order}$ as:
$$s_{order} =\textbf{W}_{order} H_{order} + \textbf{b}_{order} $$
where $\textbf{W}_{order}$ and $\textbf{b}_{order} $ are parameters for the parsing order scorer. Finally, the one-shot arc score is:
$$s_{final} = s_{arc}  + s_{order} $$

Similarly, we use the biaffine scorer for dependency label classification. We apply MLPs to the contextualized representations before using them in the label classifier as well.  As with other graph-based models, the predicted tree at training time has each word as a dependent of its highest-scoring head (although at test time we ensure that the parse is a well-formed tree via the greedy parsing algorithm).

\subsection{Training Objectives}

To parse the syntax tree $y$ for a sentence $x$ with length $l$, the easy-first model relies on an action-by-action process performed on \emph{pending}. In the training stage, the loss is accumulated once per step (action), and the model is updated by gradient backpropagation according to a preset frequency. This prohibits parallelism during model training lack between and within sentences. Therefore, the traditional easy-first model was trained to maximize following probability:
$$P_\theta(y|x)  = \prod_{i=1}^{l-1} P_\theta(y^{act}_i|\emph{pending}_i),$$
where $\emph{pending}_i$ is the pending list state at step $i$.

While for our proposed model, it uses the training method similar to that of graph-based models, in which the arc scores are all obtained in one-shot. Consequently, it does not  rely on the pending list in the training phase and only uses the pending list to promote the process of linear parsing in the inference stage. Our model is trained to optimize the probability of the dependency tree $y$ when given a sentence $x$: $P_\theta(y|x)$, which can be factorized as:
\begin{align*}
	P_\theta(y|x)  = \prod_{i=1}^l P_\theta(y^{arc}_i, y^{rel}_i, y^{order}_i|x_i)
\end{align*}
where $\theta$ represents learnable parameters, $l$ denotes the length of the processing sentence, and $y^{arc}_i$, $y^{rel}_i$ denote the highest-scoring head and dependency relation for node $x_i$. Thus, our model factors the distribution according to a bottom-up tree structure.

Corresponding to multiple objectives, several parts compose the loss of our model. The overall training loss is the sum of three objectives: 
\begin{equation*}
\mathcal{L} = \mathcal{L}^{arc} + \mathcal{L}^{rel} + \mathcal{L}^{order},
\end{equation*}
where  the loss for arc prediction $\mathcal{L}^{arc}$ is the negative log-likelihood loss of the golden structure $y^{arc}$:
\begin{equation*}
\mathcal{L}^{arc}(x) = - \log P_\theta(y^{arc}|x),
\end{equation*}
the loss for relation prediction $\mathcal{L}^{rel}$ is implemented as the negative log-likelihood loss of the golden relation $y^{rel}$ with the golden structure $y^{arc}$, 
\begin{equation*}
\mathcal{L}^{rel}(x, y^{arc}) = - \log P_\theta(y^{rel}|x, y^{arc}),
\end{equation*}
and the loss for parsing order prediction $\mathcal{L}^{order}$:
\begin{equation*}
\mathcal{L}^{order}(x) = - \log P_\theta(y^{order}|x).
\end{equation*}
Because the parsing order score of each layer in the tree increases by 1, we frame it as a classification problem and therefore add a multi-class classifier module as the order scorer.

\begin{table*}[h]
	\centering
	\scalebox{0.9}{
		\begin{tabular}{l|c|cc|cc}
			\hline
			\hline
			& & \multicolumn{2}{c|}{\textbf{PTB-SD}} & \multicolumn{2}{c}{\textbf{CTB}} \\
			\textbf{System} & \textbf{Method} & LAS(\%) & UAS(\%) & LAS(\%) & UAS(\%) \\
			\hline
			\hline
			\cite{dyer2015transition} & T (g) & 90.9 & 93.1 & 85.5 & 87.1 \\
			\cite{kiperwasser2016simple} & T (g) & 91.9 & 93.9 & 86.1 & 87.6 \\
			\cite{andor2016globally} & T (b)  & 92.79 & 94.61 & - & - \\
			\cite{zhu2015re} & T (re) & - & 94.16 & - & 87.43 \\ %
			S\text{\small{TACK}}P\text{\small{TR}}: \cite{ma2018stack} & T (g) & 94.19 & 95.87 & \textbf{89.29} & \textbf{90.59} \\
			\hline
			\cite{zhang2014enforcing} & G (3rd)& 90.64 & 93.01 & 86.34 & 87.96 \\
			\cite{zhang-etal-2016-probabilistic} & G (3rd) & 91.29 & 93.42 & 86.17 & 87.65 \\
			\cite{wang2016graph} & G (1st)& 91.82 & 94.08 & 86.23 & 87.55 \\
			\cite{kiperwasser2016simple} & G (1st)& 90.9 & 93.0 & 84.9 & 86.5 \\
			B\text{\small{I}}AF: \cite{dozat2017deep} & G (1st)& 94.08 & 95.74 & 88.23 & 89.30 \\
			B\text{\small{I}}AF [re-impl]: \cite{ma2018stack} & G (1st)& 94.21& 95.84 & 89.14 & 90.43 \\
			WO: \cite{D18-1311} & G (1st) & 94.54& 95.66 & - & - \\
			Division: \cite{zhou2019head} & G & 93.09 & 94.32 & 87.31 & 89.14  \\
			Joint: \cite{zhou2019head} & G & 94.68 & \textbf{96.09} & 89.15 & \textbf{91.21} \\
			\textbf{This work (MST)} & G (1st) & 94.57 & 95.93 & 89.45 & 90.55  \\
			\hline
			\cite{kiperwasser2016easy} & E (g)  & 90.9 & 93.0 & 85.5 & 87.1 \\
			\textbf{This work} & G + E (g) & 94.54 & 95.83 & 89.44  & 90.47  \\
			\hline
			WO + ELMo: \cite{D18-1311} & G (1st) & \textbf{95.25} & 96.35 & - & - \\
			Division + BERT: \cite{zhou2019head} & G & 94.56 & 96.22 & - & -  \\
			Joint + BERT: \cite{zhou2019head} & G & 95.43 & \textbf{97.00} & 89.15 & \textbf{91.21} \\
			\textbf{This work + ELMo} & G + E (g) & 94.57 & \textbf{96.37} & \textbf{89.45}$^*$ & 90.51$^*$ \\
			\textbf{This work + BERT} & G + E (g) & 94.63 &  \textbf{96.44} & \textbf{89.73} &  \textbf{90.89} \\
			\textbf{This work + ELMo + BERT} & G + E (g) & 94.81 &  \textbf{96.59} & \textbf{89.73}$^*$ & \textbf{90.88}$^*$  \\
			\hline
			\hline
		\end{tabular}
	}
	\caption{Comparison of results on the test sets. ``T", ``G" and ``E" indicate transition-based, graph-based and easy-first models, respectively. The ``G + E" represents the graph-based training while the easy-first algorithm is used for inference. Acronyms used: (g) -- greedy, (b) -- beam search, (re) -- re-ranking, (3rd) -- 3rd-order, (1st) -- 1st-order. The ``*" in the upper right corner of the results is because the original ELMo \cite{peters2018deep} has no Chinese version. We instead used the multilingual version ``HIT-ELMo" pre-trained by \cite{che-EtAl:2018:K18-2}.} \label{overall}
\end{table*}

\subsection{Non-Projective Inference}
For non-projective inference, we introduce two additional arc-building actions as follows.

$\bullet$ NP-\text{\large{A}}TTACH\text{\large{L}}EFT($i$): attaches $p_{j}$ to $p_i$ where $j > i$, which builds an arc ($p_i$, $p_{j}$) headed by $p_i$, and removes $p_{j}$ from \emph{pending}.

$\bullet$ NP-\text{\large{A}}TTACH\text{\large{R}}IGHT($i$): attaches $p_{j}$ to $p_i$ where $j < i$ which builds an arc ($p_i$, $p_j$) headed by $p_i$, and removes $p_j$ from \emph{pending}.



If we use the two arc-building actions for non-projective dependency trees directly on $s_{final}$, the time complexity will become $O(n^3)$, so we need to modify this algorithm to accommodate the non-projective dependency trees. Specifically, we no longer use $s_{final}$ directly for greedy search but instead divide each decision into two steps. The first step is to use the order score $s_{order}$ to sort the pending list in descending order. Then, the second step is to find the edge with the largest arc score $s_{arc}$ for this node in the first position of the pending list. 

\begin{table*}
	\centering
	\scalebox{1.0}{
	\begin{tabular}{l|c|c|c|c|c|c}
		\hline
		& Bi-Att & NeuroMST & B\text{\small{I}}AF & S\text{\small{TACK}}P\text{\small{TR}} & Ours & Best Published \\
		\hline
		& UAS [LAS] & UAS [LAS] & UAS [LAS] & UAS [LAS] & UAS [LAS] & UAS [LAS] \\
		\hline
		ar & 80.34 [68.58] & 80.80 [69.40] & 82.15 [71.32] & 83.04 [72.94] & \textbf{84.47 [74.12]} & 81.12 [-] \\
		bg & 93.96 [89.55] & 94.28 [90.60] & 94.62 [91.56] & 94.66 [91.40] & \textbf{95.63 [92.47]} & 94.02 [-] \\
		cs & 91.16 [85.14] & 91.18 [85.92] & 92.24 [87.85] & 92.83 [88.75] & \textbf{93.71 [89.90]} & 91.16 [85.14] \\
		da & 91.56 [85.53] & 91.86 [87.07] & 92.80 [88.36] & 92.08 [87.29] & \textbf{93.06 [88.45]} & 92.00 [-]\\
		de & 92.71 [89.80] &93.62 [91.90] & 94.52 [93.06] & 94.77 [93.21] &  \textbf{95.34 [94.17]} & 92.71 [89.80] \\
		en & - & 94.66 [92.52] & 95.19 [93.14] & 93.25 [93.17] & \textbf{95.48 [93.64]} & 93.25 [-]\\
		es & 88.74 [84.03] & 89.20 [85.77] & 90.43 [87.08] & 90.87 [87.80] & \textbf{91.92 [88.74]} & 88.75 [84.03]\\
		ja & 93.44 [90.67] & \textbf{94.02 [92.60]} & 93.95 [92.46] & 93.38 [91.92] & 93.70 [92.05] & 93.80 [-] \\
		nl & 87.15 [82.41] & 87.85 [84.82] & 90.07 [87.24] & 90.10 [87.05] & \textbf{91.84 [89.30]} & 87.39 [-]\\
		pt & 92.77 [88.44] &92.71 [88.92] & 93.41 [89.96] & 93.57 [90.07] & \textbf{93.85 [90.56]} & 93.03 [-]\\
		sl & 86.01 [75.90] & 86.73 [77.56] & 87.55 [78.52] & 87.59 [78.85] & \textbf{88.91 [79.82]} & 87.06 [-]\\
		sv & 90.50 [84.05]& 91.22 [86.92] & 92.22 [88.44] & 92.49 [89.01] & \textbf{93.90 [89.91]} & 91.85 [85.26]\\
		tr & 78.43 [66.16] & 77.71 [65.81] & 79.84 [68.63] & 79.56 [68.03] & \textbf{82.06 [71.50]} & 78.43 [66.16]\\
		zh & - & 93.40 [90.10]  & 94.05 [90.89] & 93.88 [90.81] & \textbf{94.87 [91.52]} & 93.04 [-] \\
		\hline
		\textbf{avg} & 88.89 [82.52] & 89.94 [84.99] & 90.93 [86.32] & 90.86 [86.45] & \textbf{92.05 [87.58]} & \\
		\hline
	\end{tabular}
	}
	\caption{UAS and LAS on 14 treebanks from CoNLL shared tasks, together with several state-of-the-art parsers. Bi-Att is the bi-directional attention based parser \cite{cheng2016bi}, and NeuroMST is the neural MST parser \cite{ma2017neural}. ``Best Published" includes the best results in recent years among \cite{koo2010dual}, \cite{martins2011dual}, \cite{martins2013turning}, \cite{lei2014low}, \cite{zhang2014greed}, \cite{zhang2014enforcing}, \cite{pitler2015linear}, and \cite{cheng2016bi}  in addition to the ones we listed above.} \label{conll_results}
\end{table*}

\subsection{Time Complexity}


The number of decoding steps to build a parse tree for a sentence is the same as its length, $n$. Combining this with the searching in the \emph{pending} list (at each step, we need to find the highest-scoring pair in the pending list to attach. This has a runtime of $O(n)$. The time complexity of a full decoding is $O(n^2)$, which is equal to 1$^{st}$-order non-projective graph-based parsing but more efficient than 1$^{st}$-order projective parsing with $O(n^3)$ and other higher order graph parsing models. Compared with the current state-of-the-art transition-based parser S\text{\small{TACK}}P\text{\small{TR}} \cite{ma2018stack}, with the same decoding time complexity as ours, since our number of decoding takes $n$ steps while S\text{\small{TACK}}P\text{\small{TR}} takes $2n-1$ steps for decoding and needs to compute the attention vector at each step, our model actually would be much faster than S\text{\small{TACK}}P\text{\small{TR}} in decoding.



For the non-projective inference in our model, the complexity is still $O(n^2)$. Since the order score and the arc score are two parts that do not affect each other, we can sort the order scores with time complexity of $O$($n$log$n$) and then iterate in this descending order. The iteration time complexity is $O(n)$ and determining the arc is also $O(n)$, so the overall time complexity is $O$($n$log$n$) $+$ $O(n^2)$, simplifying to $O(n^2)$.

\section{Experiments}
We evaluate our parsing model on the English Penn Treebank (PTB), the Chinese Penn Treebank (CTB), treebanks from two CoNLL shared tasks and the Universal Dependency (UD) Treebanks, using unlabeled attachment scores (UAS) and labeled attachment scores (LAS) as the metrics. Punctuation is ignored as in previous work \cite{dozat2017deep}.  
For English and Chinese, we use the projective inference, while for other languages, we use the non-projective one.

\subsection{Treebanks} 
For English, we use the Stanford Dependency (SD 3.3.0) \cite{de2008stanford} conversion of the Penn Treebank \cite{marcus1993building}, and follow the standard splitting convention for PTB, using sections 2-21 for training, section 22 as a development set and section 23 as a test set. We use the Stanford POS tagger \cite{toutanova2003feature} generate predicted POS tags.

For Chinese, we adopt the splitting convention for CTB \cite{xue2005the} described in \cite{dyer2015transition}. The dependencies are converted with the Penn2Malt converter. Gold segmentation and POS tags are used as in previous work \cite{dyer2015transition}.

For the CoNLL Treebanks, we use the English treebank from the CoNLL-2008 shared task \cite{surdeanu2008conll} and all 13 treebanks from the CoNLL-X shared task \cite{buchholz2006conll}. The experimental settings are the same as \cite{ma2015efficient}.

For UD Treebanks, following the selection of \cite{ma2018stack}, we take 12 treebanks from UD version 2.1 (Nivre et al. 2017): Bulgarian (bg), Catalan (ca), Czech (cs), Dutch (nl), English (en), French (fr), German (de), Italian (it), Norwegian (no), Romanian (ro), Russian (ru) and Spanish (es). We adopt the standard training/dev/test splits and use the universal POS tags provided in each treebank for all the languages.


\subsection{Implementation Details}

\paragraph{Pre-trained Embeddings} We use the GloVe \cite{pennington2014glove} trained on Wikipedia and Gigaword as external embeddings for English parsing. For other languages, we use the word vectors from 157 languages trained on Wikipedia and Crawl using fastText \cite{grave2018learning}. We use the extracted BERT layer features to enhance the performance on CoNLL-X and UD treebanks.

\paragraph{Hyperparameters} The character embeddings are 8-dimensional and randomly initialized. In the character CNN, the convolutions have a window size of 3 and consist of 50 filters. We use 3 stacked bidirectional LSTMs with 512-dimensional hidden states each. The outputs of the BiLSTM employ a 512-dimensional MLP layer for the arc scorer, a 128-dimensional MLP layer for the relation scorer, and a 128-dimensional MLP layer for the parsing order scorer, with all using  ReLU as the activation function. Additionally, for parsing the order score, since considering it a classification problem over parse tree layers, we set its range\footnote{Scores above 32 will be truncated to 32.} to $[0, 1,  ..., 32]$.

\paragraph{Training} Parameter optimization is performed with the Adam optimizer with $\beta_1$ = $\beta_2$ = 0.9. We choose an initial learning rate of $\eta_0$ = 0.001. The learning rate $\eta$ is annealed by multiplying a fixed decay rate $\rho$ = 0.75 when parsing performance stops increasing on validation sets. To reduce the effects of an exploding gradient, we use a gradient clipping of 5.0. For the BiLSTM, we use recurrent dropout with a drop rate of 0.33 between hidden states and 0.33 between layers. Following \cite{dozat2017deep}, we also use embedding dropout with a rate of 0.33 on all word, character, and POS tag embeddings. 

\begin{table}
	\centering
	\setlength{\tabcolsep}{1mm}
	\begin{tabular}{l|c|c|c}
		\hline
		& B\text{\small{I}}AF & S\text{\small{TACK}}P\text{\small{TR}} & Ours \\
		\hline
		& UAS [LAS] & UAS [LAS] & UAS [LAS] \\
		\hline
		bg & 94.30 [90.04] & 94.31 [89.96] & \textbf{95.43 [91.27]} \\
		ca & 94.36 [92.05] & 94.47  [92.39] & \textbf{95.38 [93.57]} \\
		cs & 94.06 [90.60]   & 94.21 [90.94] & \textbf{95.08 [91.52]} \\
		de & 90.26 [86.11] & 90.26   [86.16]& \textbf{91.33 [87.27]}  \\
		en & 91.91 [89.82]  & 91.93  [89.83] & \textbf{93.17 [90.79]} \\
		es & 93.72 [91.33] & 93.77  [91.52] & \textbf{94.83 [92.36]} \\
		fr & 92.62 [89.51] & 92.90  [89.88]& \textbf{93.95 [91.42]} \\
		it & 94.75 [92.72] & 94.70  [92.55] & \textbf{95.73 [93.52]} \\
		nl & 93.44 [91.04] & 93.98  [91.73] & \textbf{95.32 [92.82]} \\
		no & 95.28 [93.58] & 95.33  [93.62] & \textbf{95.87 [94.15]} \\
		ro & 91.94 [85.61]  & 91.80  [85.34] & \textbf{92.72 [86.16]} \\
		ru & 94.40 [92.68] & 94.69 [93.07] & \textbf{95.88 [94.26]}\\
		\hline
		\textbf{avg} & 93.42 [90.42]  & 93.52 [90.66] & \textbf{94.55 [91.59]} \\
		\hline
	\end{tabular}
	\caption{UAS and LAS on test datasets of 12 treebanks from UD Treebanks, together with B\text{\small{I}}AF and S\text{\small{TACK}}P\text{\small{TR}} for comparison.} \label{ud_results}
\end{table}

\subsection{Main Results}

We now compare our model with several other recently proposed parsers as shown in Table \ref{overall}. Our global greedy parser significantly outperforms the easy-first parser in \cite{kiperwasser2016easy} (HT-LSTM) on both PTB and CTB. Compared with other graph- and transition-based parsers, our model is also competitive with the state-of-the-art on PTB when considering the UAS metric. Compared to state-of-the-art parsers in transition and graph types, B\text{\small{I}}AF and S\text{\small{TACK}}P\text{\small{TR}}, respectively, our model gives better or comparable results but with much faster training and decoding. Additionally, with the help of pre-trained language models, ELMo or BERT, our model can achieve even greater results. 

In order to explore the impact of the parsing order objective on the parsing performance, we replace the greedy inference with the traditional MST parsing algorithm (i.e., B\text{\small{I}}AF + parsing order objective), and the result is shown as ``This work (MST)", giving slight performance improvement compared to the greedy inference, which shows globally optimized decoding of graph model still takes its advantage. Besides, compared to the standard training objective for graph model based parser, the performance improvement is slight but still shows the proposed parsing order objective is indeed helpful.


%


\subsection{CoNLL Results}

Table \ref{conll_results} presents the results on 14 treebanks from the CoNLL shared tasks. Our model yields the best results on both UAS and LAS metrics of all languages except the Japanese. As for Japanese, our model gives unsatisfactory results because the original treebank was written in Roman phonetic characters instead of hiragana, which is used by both common Japanese writing and our pre-trained embeddings. Despite this, our model overall still gives 1.0\% higher average UAS and LAS than the previous best parser, B\text{\small{I}}AF.

\subsection{UD Results}

Following \cite{ma2018stack}, we report results on the test sets of 12 different languages from the UD treebanks along with the current state-of-the-art: B\text{\small{I}}AF and S\text{\small{TACK}}P\text{\small{TR}}. Although both B\text{\small{I}}AF and S\text{\small{TACK}}P\text{\small{TR}} parsers have achieved relatively high parsing accuracies on the 12 languages and have all UAS higher than 90\%, our model achieves state-of-the-art results in all languages for both UAS and LAS. Overall, our model reports more than 1.0\% higher average UAS than S\text{\small{TACK}}P\text{\small{TR}} and 0.3\% higher than B\text{\small{I}}AF.


\subsection{Runtime Analysis}

\begin{table}
	\centering
	\small
	\begin{tabular}{l|c|c}
		\toprule
		\multirow{2}{*}{Systems}& Training &  Decoding \\
		& T (Hours) & S (tokens/sec.) \\
		\midrule
		Origin Easy-First & $\approx$16  &  8532 \\
		B\text{\small{I}}AF & $\approx$8  &  2568 \\
		S\text{\small{TACK}}P\text{\small{TR}} & $\approx$12   &  1508 \\
		\textbf{Ours} & $\approx$8 &  8541 \\
		\bottomrule
	\end{tabular}
	\caption{Training time and decoding speed. The experimental environment is on the same machine with Intel i9 9900k CPU and NVIDIA 1080Ti GPU.} \label{time}
\end{table}

In order to verify the time complexity analysis of our model, we measured the running time and speed of B\text{\small{I}}AF, S\text{\small{TACK}}P\text{\small{TR}} and our model on PTB training and development set using the projective algorithm. The comparison in Table \ref{time} shows that in terms of convergence time, our model is basically the same speed as B\text{\small{I}}AF, while S\text{\small{TACK}}P\text{\small{TR}} is much slower. For decoding, our model is the fastest, followed by B\text{\small{I}}AF.  S\text{\small{TACK}}P\text{\small{TR}} is unexpectedly the slowest. This is because the time cost of attention scoring in decoding is not negligible when compared with the processing speed and actually even accounts for a significant portion of the runtime.

\subsection{Results under Pretrain-finetune Pattern}
Using the fixed representation from pre-trained language models has led to significant performance gains. However, the pre-trained language models may be more flexibly used in a fine-tuning way rather than only initializing the embeddings from the models.We consider three baseline parsers for the fine-tuning usage of the pre-trained language models (FT-PLM). The first is the B\text{\small{I}}AF baseline, which uses MST as the inference algorithm; the second is the proposed global greedy parser, which uses the proposed global greedy projective inference algorithm; the third is the global greedy parser with the non-projective inference algorithm. The experiment is based on PTB, and for the pre-trained language model, we select 6 mainstream pretrained models: BERT \cite{devlin2018bert} (\textbf{bert-large-cased}), XLNet \cite{yang2019xlnet} (\textbf{xlnet-large-cased}), RoBERTa \cite{liu2019roberta} (\textbf{roberta-large}), XLM-RoBERTa \cite{conneau2019unsupervised} (\textbf{xlm-roberta-large}), DistilBERT \cite{sanh2019distilbert} (\textbf{distilbert-base-uncased}), ALBERT \cite{lan2019albert} (\textbf{albert-xxlarge-v1},\textbf{albert-xxlarge-v2}). In addition, we also report the accuracy (Order Acc.) of the parsing order module in our proposed global greedy model which shows that the ratio of the node parsing order prediction by the parsing order module to that of the golden tree.

The results are shown in Tables \ref{pf_baseline}, \ref{our_ggp} and \ref{our_ggnp}. Table \ref{pf_baseline} shows the results of our re-implementation of the FT-PLM-based B\text{\small{I}}AF baseline. It can be seen from the results that we have provided several strong baselines for the dependency parsing field only with the help of pre-trained language model and the \textbf{xlnet-large-cased} achieved the state-of-the-art parsing performance, which verifies that pre-trained language model is helpful in improving parsing performance.

Tables \ref{our_ggp} and \ref{our_ggnp} show the performance of our global greedy model under different inference algorithms. The result of our proposed global greedy model is slightly lower than that of baseline, but the time complexity of our inference algorithm is lower than that of baseline.

\begin{table}
	\centering
	\small
	\begin{tabular}{l|c|c}
		\toprule
		Systems & UAS &  LAS \\
		\midrule
		\textbf{bert-large-cased}$^{*}$ & 96.55 &	94.70 \\
		\midrule
		\textbf{bert-large-cased} & 96.92 &	95.35 \\
		\textbf{xlnet-large-cased} & \textbf{97.22} & \textbf{95.66} \\
		\textbf{roberta-large} & 97.11 &	95.51 \\
		\textbf{xlm-roberta-large} & 97.06 & 95.56 \\
		\textbf{distilbert-base-uncased} & 96.13 & 94.43 \\
		\textbf{albert-xxlarge-v1} & 97.09 & 95.24 \\
		\textbf{albert-xxlarge-v2} & 97.17 & 95.45 \\
		\bottomrule
	\end{tabular}
	\caption{The results of FT-PLM-based B\text{\small{I}}AF baseline with MST inference algorithm, $^{*}$ indicate the results of model using fixed pre-trained language model representation.} \label{pf_baseline}
\end{table}

\begin{table}
	\centering
	\small
	\begin{tabular}{l|c|c|c}
		\toprule
		Systems & UAS &  LAS & Order Acc. \\
		\midrule
		\textbf{bert-large-cased}$^{*}$ & 96.44 & 94.63 & 90.25 \\
		\midrule
		\textbf{bert-large-cased} & 96.57 & 95.05 & 90.36 \\
		\textbf{xlnet-large-cased} & \textbf{96.97} & \textbf{95.37} & \textbf{91.59}\\
		\textbf{roberta-large} & 96.86 & 95.27 & 91.34 \\
		\textbf{xlm-roberta-large} & 96.74 & 95.15 & 91.03 \\
		\textbf{albert-xxlarge-v1} & 96.69 & 94.89 & 91.57 \\
		\textbf{albert-xxlarge-v2} & 96.75 & 94.93 & 91.11 \\
		\bottomrule
	\end{tabular}
	\caption{The results of our FT-PLM-based model with global greedy projective inference algorithm, $^{*}$ indicate the results of model using fixed pre-trained language model representation.} \label{our_ggp}
\end{table}

\begin{table}[h]
	\centering
	\small
	\begin{tabular}{l|c|c|c}
		\toprule
		Systems & UAS &  LAS & Order Acc. \\
		\midrule
		\textbf{bert-large-cased}$^{*}$ & 96.51 & 94.77 & 90.25 \\
		\midrule
		\textbf{bert-large-cased} & 96.70 & 95.20 & 90.36 \\
		\textbf{xlnet-large-cased} & \textbf{97.12} & \textbf{95.53} & \textbf{91.59} \\
		\textbf{roberta-large} & 96.99 & 95.37 & 90.92 \\
		\textbf{xlm-roberta-large} & 96.98 & 95.38 & 91.16 \\
		\textbf{albert-xxlarge-v1} & 96.79 & 94.97 & 91.57 \\
		\textbf{albert-xxlarge-v2} & 96.96 & 95.13 & 91.11 \\
		\bottomrule
	\end{tabular}
	\caption{The results of our FT-PLM-based model with global greedy non-projective inference algorithm, $^{*}$ indicate the results of model using fixed pre-trained language model representation.} \label{our_ggnp}
\end{table}

\section{Conclusion}
This paper presents a new global greedy parser in which we enable greedy parsing inference compatible with the global arc scoring of graph-based parsing models instead of the local feature scoring of transitional parsing models. The proposed parser can perform projective parsing when only using two arc-building actions, and it also supports non-projective parsing when introducing two extra non-projective arc-building actions. Compared to graph-based and transition-based parsers, our parser achieves a better tradeoff between parsing accuracy and efficiency by taking advantages of both graph-based models' training methods and transition-based models' linear time decoding strategies. Experimental results on 28 treebanks show the effectiveness of our parser by achieving good performance on 27 treebanks, including the PTB and CTB benchmarks.

\bibliographystyle{aaai}
\bibliography{references}

\end{document}